\documentclass[10pt,twocolumn,letterpaper]{article}

\usepackage{makecell}
\newcommand{\PreserveBackslash}[1]{\let\temp=\\#1\let\\=\temp}
\newcolumntype{C}[1]{>{\PreserveBackslash\centering}p{#1}}
\newcolumntype{R}[1]{>{\PreserveBackslash\raggedleft}p{#1}}
\newcolumntype{L}[1]{>{\PreserveBackslash\raggedright}p{#1}}

\usepackage{cvpr}
\usepackage{times}
\usepackage{epsfig}
\usepackage{graphicx}
\usepackage{amsmath}
\usepackage{amssymb}
\usepackage{mathrsfs}
\usepackage{multirow}
\usepackage[labelsep=period]{caption}
\usepackage{subfig}

\usepackage[breaklinks=true,bookmarks=false]{hyperref}

\cvprfinalcopy 

\def\cvprPaperID{4396} 

\ifcvprfinal\fi
\begin{document}
\title{Interaction-and-Aggregation Network for Person Re-identification}

\author{Ruibing Hou$^{1,2}$, Bingpeng Ma$^{2}$, Hong Chang$^{1,2}$, Xinqian Gu$^{1,2}$, Shiguang Shan$^{1,2,3}$, Xilin Chen$^{1,2}$\\
$^1$Key Laboratory of Intelligent Information Processing of Chinese Academy of Sciences (CAS),\\Institute of Computing Technology, CAS, Beijing, 100190, China\\
$^2$University of Chinese Academy of Sciences, Beijing, 100049, China\\
$^3$CAS Center for Excellence in Brain Science and Intelligence Technology, Shanghai, 200031, China\\
{\tt\small \{ruibing.hou, xinqian.gu\}@vipl.ict.ac.cn, bpma@ucas.ac.cn, \{changhong, sgshan,xlchen\}@ict.ac.cn}
}

\maketitle
\thispagestyle{empty} 

\begin{abstract}
Person re-identification (reID) benefits greatly from deep convolutional neural networks (CNNs) which learn robust feature embeddings. However, CNNs are inherently limited in modeling the large variations in person pose and scale due to their fixed geometric structures. In this paper, we propose a novel network structure, \textit{Interaction-and-Aggregation} (IA), to enhance the feature representation capability of CNNs. Firstly, \textit{Spatial IA} (SIA) module is introduced. It models the interdependencies between spatial features and then aggregates the correlated features corresponding to the same body parts. Unlike CNNs which extract features from fixed rectangle regions, SIA can adaptively determine the receptive fields according to the input person pose and scale. Secondly, we introduce \textit{Channel IA} (CIA) module which selectively aggregates channel features to enhance the feature representation, especially for small-scale visual cues. Further, IA network can be constructed by inserting IA blocks into CNNs at any depth. We validate the effectiveness of our model for person reID by demonstrating its superiority over state-of-the-art methods on three benchmark datasets.
\end{abstract}

\section{Introduction}
Person re-identification (reID) aims at identifying a person of interest across different cameras with a given probe. It plays a significant role in intelligent surveillance systems. In recent years, Deep Convolutional Neural Networks (CNNs), which typically stack convolution and pooling layers to learn discriminative features, have obtained state-of-the-art results for person reID. Despite of years of efforts, there still exist many challenges such as large variations in person pose, scale, and background clutter.

\begin{figure}[t]
\centering
   \includegraphics[width=0.9\linewidth]{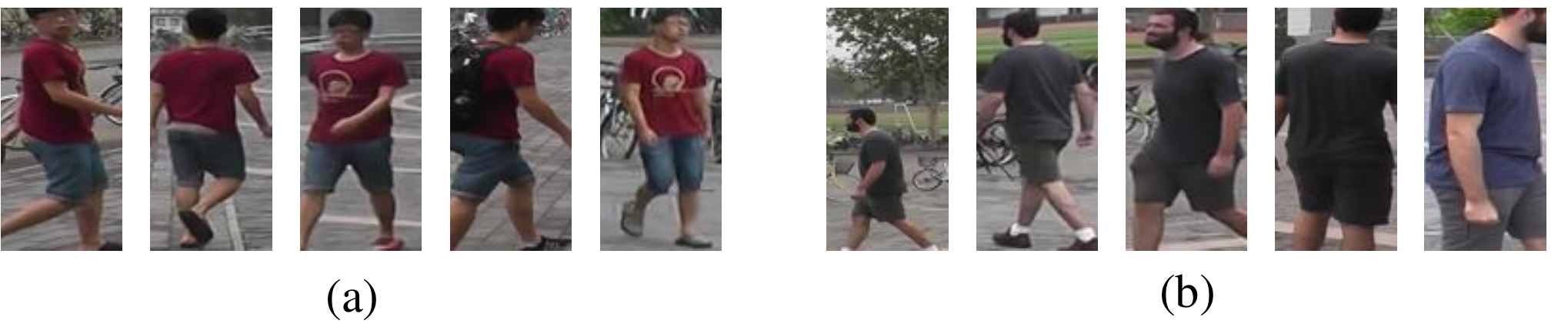}
   \captionsetup{font={small}}
   \caption{The critical influencing factors for person reID. (a) A person appears in various poses; (b) various scales due to imperfect detection results.}
\label{fig1}
\end{figure}

Body part misalignment is a critical influencing factor on reID results, which can be attributed to two causes. First, pedestrians naturally take on various poses as shown in Fig.~\ref{fig1} (a). Second, the body parts have various scales across different images of the same person caused by imperfect pedestrian detection, as illustrated in Fig.~\ref{fig1} (b). To resolve these problems, some approaches have been proposed recently. One way is to localize body parts explicitly and combine the representations over them~\cite{spindle-net,pose-invariant,pose-driven,mask-guided,semantic}. This scheme requires highly-accurate part detection. Unfortunately, even state-of-the-art part detection solutions are not perfect. Another type of methods resorts to multi-scale features fusion where the feature maps are computed at multiple layers of a network~\cite{CRF,Resource,Multi-Level,Efficient}. Nevertheless, these methods only employ manually specified scales, which are ineffective to model large scale variations. In short, existing methods, which attempt to utilize body part detection or multi-scale features, are still limited in modeling the large variations in body pose and scale.

An essential reason why these approaches are not robust to body pose and scale variations is that they all use CNNs to extract pedestrian features. Actually, CNNs are inherently limited in modeling large geometric transformations. The limitation originates from the fixed geometric structures of CNNs modules: a convolution unit which samples the input feature map at fixed locations and a pooling layer which reduces the spatial resolution at a fixed ratio. There lacks internal mechanisms to handle the body pose and scale variations. \textit{For one thing}, the receptive fields of the feature maps are pre-defined rectangles, which can not adaptively localize the non-rigid body parts with different poses. \textit{For another}, the receptive fields of all activation units in the same CNN layer have the same size, which is undesirable for high level CNN layers to encode semantics for body parts of different scales.

In this paper, we propose a new network structure, \textit{Interaction-and-Aggregation} (IA), to enhance the feature representation capability of CNNs, especially at the presence of body pose and scale variations.
IA consists of two modules: \textit{Spatial Interaction-and-Aggregation} (SIA) and \textit{Channel Interaction-and-Aggregation} (CIA).
Unlike CNNs which extract features with fixed geometric structure, SIA adaptively determines the receptive fields according to the pose and scale of input person image. More specifically, given the intermediate feature maps from CNNs, SIA generates spatial semantic relation maps to discover two types of interdependencies between different image positions: appearance relations where positions with similar feature representations have a higher correlation, and location relations where positions close to each other tend to have a higher correlation. In this way, the body parts with various poses and scales can be adaptively localized. Based on the spatial relation maps, an \textit{aggregation} operation is adopted to update the feature maps via aggregating the semantically correlated features across different positions. Similar with SIA in principle, we propose CIA to further enhance the representation power of CNNs. Unlike CNNs where the features from different channels are assumed independently, CIA explicitly models the semantic interdependencies between channels. Specially, for small-scale visual cues (\eg bags) that easily fade away in the high-level features from CNNs, CIA can selectively aggregate the semantically similar features of the visual cues across all channels to manifest their feature representations.

Both modules are computationally lightweight and impose only a slight increase in model complexity. They can be readily inserted into deep CNNs at any depth. In our work, we add IA blocks to ResNet-50~\cite{residual} to generate Interaction-and-Aggregation network (IANet) for person reID. We demonstrate the effectiveness of IANet on three reID datasets, and our method outperforms state-of-the-art methods under multiple evaluation metrics.

\section{Related Work}
\noindent \textbf{Person re-identification.}
Person reID methods focus on two key points: learning a powerful feature representation for images~\cite{multi-channel,gated,lstm,Mancs,context-aware,harmoniou,Hydraplus-net,background,smoothed} and designing an effective distance metric~\cite{Learning,stepwise,Cross-view,One-shot,hard-aware}. Recently, deep learning approaches have obtained state-of-art results for reID. We focus our discussion on  those which attempt to address the problem of body pose and scale variations.

Body part detection results have been exploited for reID to extract features robust to pose and scale variations. Most approaches attempt to localize body parts explicitly and combine the representations over global features. 
Specifically, Zhao \etal~\cite{spindle-net} used a region proposal network, which is trained on an auxiliary pose dataset, to detect body parts. Su~\etal \cite{pose-driven} proposed a sub-network to estimate the human pose that is used to crop the body parts. Besides, human parsing method~\cite{semantic,mask-guided} and body part specific attention modeling~\cite{part-aligned} had also been adopted to explicitly alleviate the pose variations problem. However, part detection in low resolution pedestrian images has its own challenges, and the inevitable detection errors could propagate to the subsequent reID task.

Another line of approaches attempts to utilize multi-scale features. Liu \etal~\cite{triplet-cnn} and Chen \etal~\cite{multi-scale} proposed an architecture consisting of multiple branches for learning multi-scale features and one branch for feature fusion. Chen \etal~\cite{CRF} and Shen \etal~\cite{KPM} used the hourglass-like network~\cite{hourglass} to generate multi-scale features. Wang \etal~\cite{Resource} and Chang \etal~\cite{Multi-Level} directly fused the feature maps across multiple layers to generate a single feature. Nevertheless, these methods employ pre-defined scales that are limited in modeling large scale variations.

In contrast to the above works that rely on part detection or pre-defined scales, our proposed SIA can adaptively localize the body parts under various poses and scales and aggregate semantic features therein. Therefore, SIA can be easily inserted into existing networks, enhancing their feature representation power.  

\noindent \textbf{Modeling geometric variations.} 
There are some works which enhance the feature representation power with respect to geometric variations. 
Traditional methods include scale invariant feature transform (SIFT)~\cite{SIFT} and ORB~\cite{orb}. A lot of recent works are aimed at CNNs. 
Some works learn invariant CNN representations with respect to specific transformations such as symmetry~\cite{symmetry}, scale~\cite{scale-invariant} and rotation~\cite{Harmonic}. However, these works assume the transformations are fixed and known, which restricts their generalization to new tasks with unknown transformations. Other works adaptively learn the spatial transformations from data. Spatial Transform Network~\cite{STN} warped the feature map via a global parametric transformation.  The works~\cite{Active-convolution,Deformable} augmented the sampling locations in the convolution with offsets and learn the offsets via back-propagation end-to-end.

Our work is fundamentally different from those works in two folds. First, the basic idea and formulation are different. The above works usually learn a parametric transformation with large amount of training data, which is infeasible for reID task with a small dataset. Differently, our proposed SIA computes spatial semantic similarities to adaptively aggregate features from same body parts without any parameters. Second, all above works do not take the channel relations into consideration. In contrast, our proposed CIA explicitly models the correlations between channels, which significantly enhances the feature representation power.

\section{Interaction-and-Aggregation Network}
In this section, we first introduce SIA and CIA modules, respectively. Then, IA block, which integrates SIA and CIA modules, is illustrated, followed by IANet for person reID. Finally, we provide some discussions on the relationships between the proposed modules and other related models.

\subsection{SIA Module}
With fixed local receptive fields, CNNs are limited in representing person images with large variations in body pose and scale. To address this problem, we design the SIA module to model spatial features interdependencies. SIA could adaptively determine the receptive field for each spatial feature, thus improving the feature robustness to body pose and scale variations.

\begin{figure}[t]
\captionsetup{font={small}}
\centering
   \includegraphics[width=1.0\linewidth]{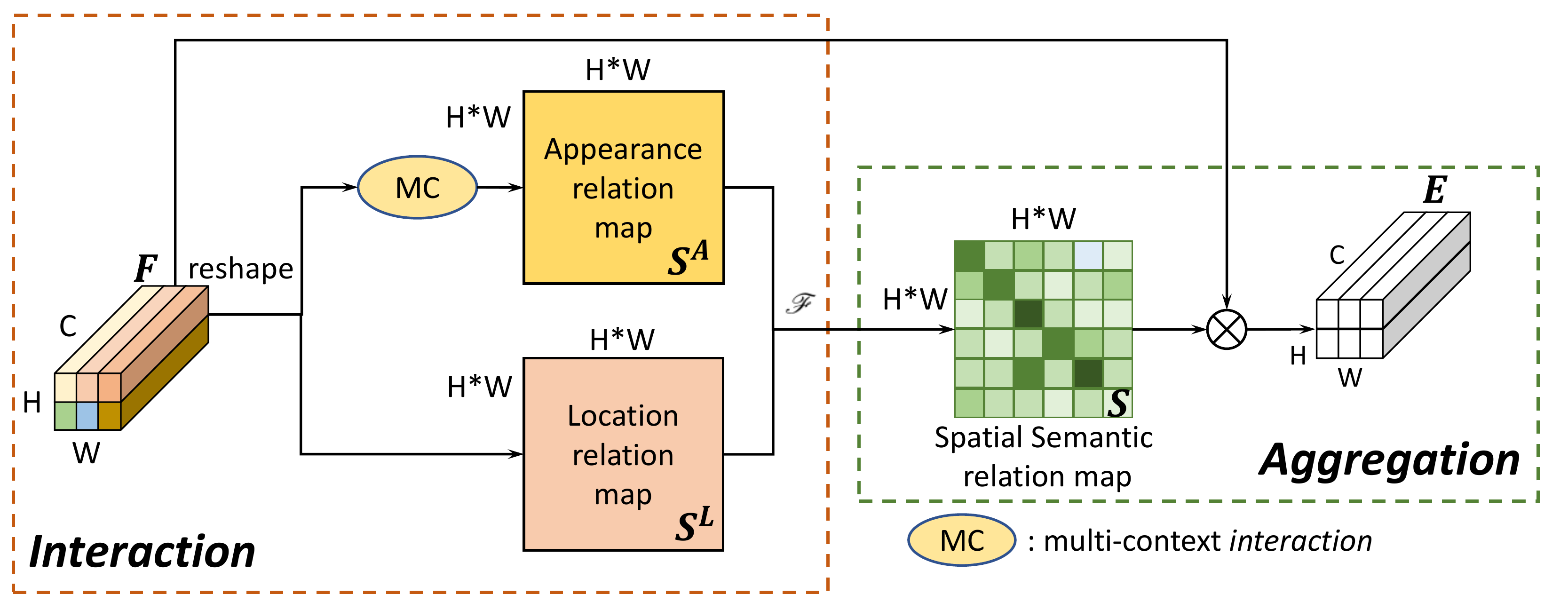}
   \caption{The architecture of Spatial \textit{Interaction-and-Aggregation} (SIA) module. We omit the softmax layer for clarity.}
\label{SIA}
\vspace*{-1em}
\end{figure}

As shown in Fig.~\ref{SIA}, suppose a convolutional feature map $F \in \mathbb{R}^{C \times H \times W}$  is given, where $C,H$ and $W$ denote the number of channels, the height and the width of the feature map respectively. We first reshape $F$ to $\mathbb{R}^{C \times M}$ where $M$ ($M=H\times W$) is the number of spatial features, then feed it into two sequential operations, \textit{interaction} and \textit{aggregation}. \textit{Interaction} operation explicitly models the interdependencies between spatial features to generate a semantic relation map $S$. Two types of relations are considered: appearance relations and location relations. 
The generated relation map is then used to aggregate correlated spatial features in the following \textit{aggregation} operation.

\begin{figure}[!t]
\centering
   \includegraphics[width=1.0\linewidth]{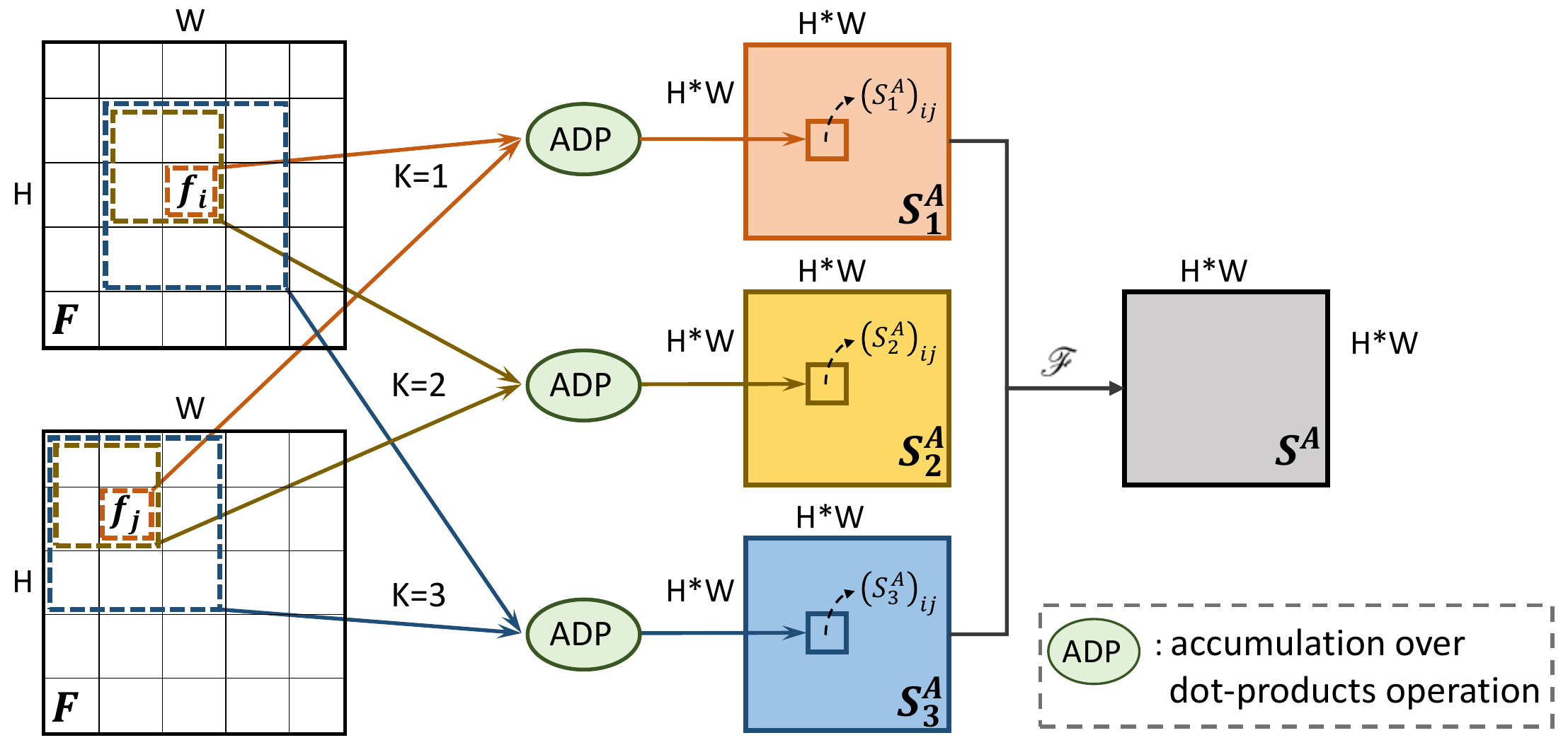}
   \captionsetup{font={small}}
   \caption{The multi-context \textit{interaction} operation of SIA. For clarity, we omit the channel dimensions of the input feature map and the softmax layer. The number of context levels is $3$ in this figure.}
\label{MS-SIA}
\vspace*{-1em}
\end{figure}

\begin{figure}[t]
\captionsetup{font={small}}
\centering
   \includegraphics[width=1.0\linewidth]{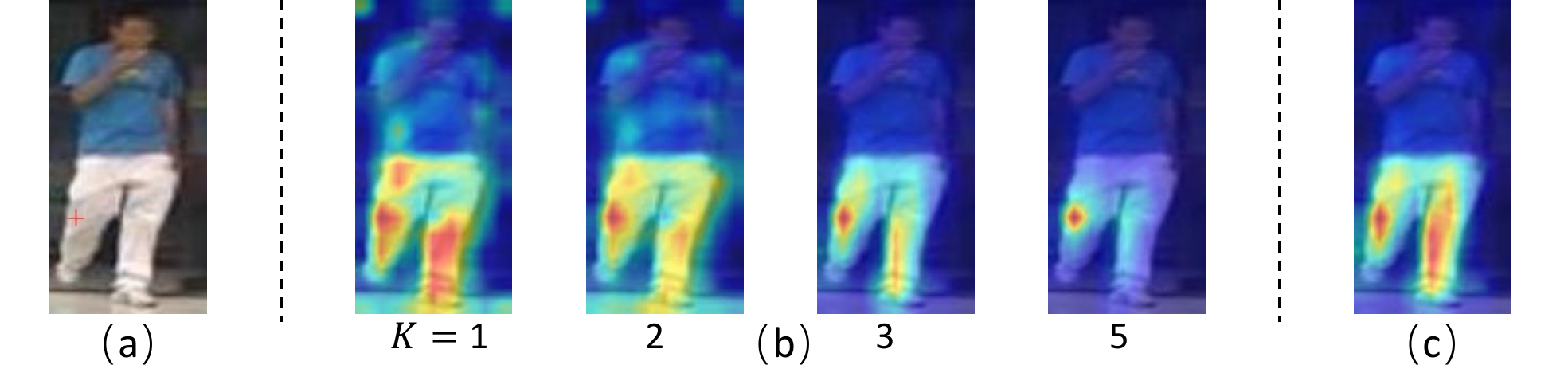}
   \caption{Visualization of the receptive fields in SIA with single-context and multi-context \textit{interaction} operations. (a) The input image, (b) The respective fields of the point marked in the input image with different single-context \textit{interaction} operations, (c) The respective fields with multi-context \textit{interaction}. Warmer color indicates higher value.}
\label{MS-Att}
\vspace{-3.5mm}
\end{figure}

\textbf{Appearance Relations.}
We measure the appearance similarity between any two positions of an input feature map to generate the appearance relation map. Du \etal~\cite{Interaction-aware} have pointed out that local features at neighboring spatial positions have high correlation since their receptive fields are often overlapped. So the patches involving  neighboring positions could capture more precise appearance. Inspired by their views, we propose to incorporate contextual information for any position in order to obtain more precise appearance similarities.
 
As illustrated in Fig.~\ref{MS-SIA}, $f_{i}$, $f_{j} \in \mathbb{R}^{C}$ denote the features in the $i^{th}$ and $j^{th}$ spatial positions of the feature map $F$. In order to calculate the appearance similarity between $f_{i}$ and $f_{j}$, we first extract the $K\times K$ patches $P_{i}$ and $P_{j}$ around $i$ and $j$, respectively. Then, the appearance similarity is obtained by accumulating the dot-products between features of corresponding positions, and then normalizing across all spatial positions in $F$ with softmax:
\vspace*{-0.1em}
\begin{equation}
\label{eq1}
\left(S^A_K\right)_{ij} = \frac{\exp\left(\sum_{k=1}^{K\times K}(p_{i,k}^T p_{j,k})\right)}{\sum_{t=1}^{H\times W} \exp\left(\sum_{k=1}^{K\times K}(p_{i,k}^T p_{t,k})\right)},
\vspace*{-0.1em}
\end{equation} where 
$p_{i,k}$ and $p_{j,k}$ denote the features in the $k^{th}$ spatial position of patches $P_{i}$ and $P_{j}$, respectively. Notably, the softmax dramatically suppresses small similarity values corresponding to different body parts. Through incorporating context and suppressing dissimilarities, the relation map can roughly localize the body parts under various poses and scales. We call the process single-context \textit{interaction} as only one patch size is considered, and $S^A_K$ is the single-context appearance relation map.

As shown in Fig.~\ref{MS-Att}, the relation maps with small context patches (\eg, $K=1$) capture more positive regions, but introduce some outliers, \eg, the located regions of the foot contain some positions corresponding to the trunk. The relation maps with large context patches (\eg, $K=5$) filter out the outliers, but ignore some positive regions. Therefore, we introduce multi-context \textit{interaction} by fusing multiple single-context relation maps with different context patch sizes. The multi-context appearance relation map $S^A$ is computed as:
\vspace*{-0.2em}
\begin{equation}
\label{eq2}
S^A = softmax\left(\mathscr{F}\left(S^A_1,\ldots,S^A_N\right)\right),
\end{equation}where $N$ denotes the number of context levels and $\mathscr{F}$ is a fusion function with element-wise product. From Fig.~\ref{MS-Att} (c), multi-context \textit{interaction} can alleviate both problems and localize the body parts more precisely.

\textbf{Location Relations.}
As for pedestrian images, local features corresponding to the same body part are spatially close. To take advantage of the spatial structure information, we introduce location relations, in which features from nearby locations have a higher correlation.

Formally, the location relation between spatial features $f_{i}$ and $f_{j}$ is computed via a two-dimensional Gaussian function as follows:
\begin{small}
\vspace*{-0.1em}
\begin{equation}
l_{ij}=\frac{1}{2\pi \sigma_{1} \sigma_{2}} \exp\left[-\frac{1}{2}\left(\frac{\left(x_j-x_i\right)^2}{\sigma_{1}^2}+\frac{\left(y_j-y_i\right)^2}{\sigma_{2}^2}\right)\right],
\label{eq3}
\end{equation} 
\end{small} where
$\left(x_{i}, y_{i}\right)$ and $\left(x_{j}, y_{j}\right)$ denote the location coordinates of features $f_{i}$ and $f_{j}$ respectively, and $\left(\sigma_1, \sigma_2\right)$ are the standard deviations used to tune the Gaussian function. We then normalize $l_{ij}$'s so that the sum of the location relation values connected to $f_i$ equals to $1$. The resulting spatial location relation map $S^L$ is:
\vspace*{-0.1em}
\begin{equation}
\label{eq4}
\left(S^L\right)_{ij} = \frac{l_{ij}}{\sum_{t=1}^{H\times W} l_{it}}.
\vspace*{-0.1em}
\end{equation}
We can see that the location relation between $f_i$ and $f_j$ exponentially decreases with the increase of their spatial distance. Notably, $S^L$ is computed based on the spatial structure of the input image, which can constrain and complement the appearance relations. 

The spatial \textbf{semantic relations} ($S$) integrates the appearance with location relations, which is formulated as:
\vspace*{-0.2em}
\begin{equation}
\label{eq5}
S = softmax\left(\mathscr{F}\left(S^A, S^L\right)\right)
\vspace*{-0.2em}
\end{equation}

\textbf{\textit{Aggregation} Operation.}
To make use of the semantic relation map in the \textit{interaction} operation, we follow it with the \textit{aggregation} operation which aims to aggregate the input spatial features based on the semantic relation map.  As shown in Fig.~\ref{SIA}, we compute the aggregated feature map $E^S \in \mathbb{R}^{C\times M}$ through matrix multiplication between $F$ and the transpose of $S$: 
\vspace*{-0.2em}
\begin{equation}
\label{eq6}
E^S = F S^T.
\vspace*{-0.2em}
\end{equation}
$E^S$ is then reshaped to $\mathbb{R}^{C\times H\times W}$ to maintain the input size. 

\begin{figure}[t]
\captionsetup{font={small}}
\centering
   \includegraphics[width=1.0\linewidth]{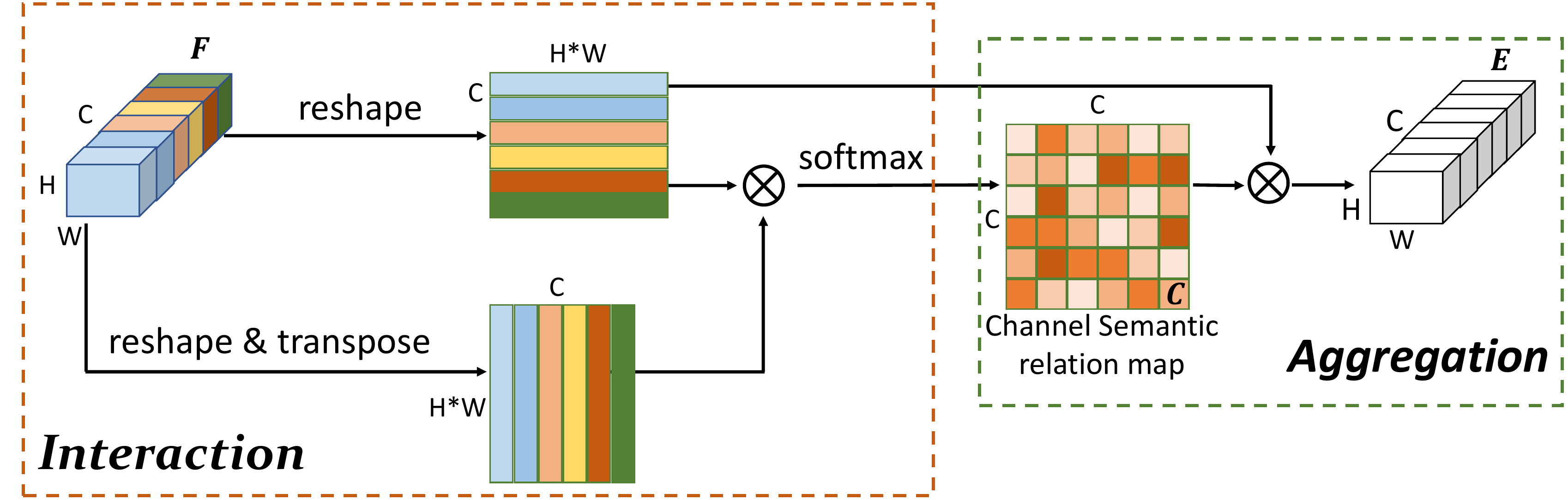}
   \caption{The architecture of Channel Interaction-and-Aggregation (CIA) module.}
\label{CIA-v1}
\vspace*{-1em}
\end{figure}

\subsection{CIA Module}
Current reID models typically stack multiple convolution layers to extract pedestrian features. With increasing the number of layers, these models could easily lose small scale visual cues, such as bags and shoes. However, these fine-grained cues are very useful to distinguish the pedestrian pairs with small inter-class variations. Zhang \etal~\cite{Occluded} have discovered that most channel maps of high-level features show strong responses for specific parts. Motivated by their views, we build the CIA module to aggregate semantically similar features across all channels, which could enhance the feature representation of specific parts. 

The structure of CIA is illustrated Fig.~\ref{CIA-v1}. In the \textit{interaction} stage, given an input convolutional feature map $F$, CIA explicitly models the semantic interdependencies between different channels of $F$ to generate a channel semantic relation map. To this end, we first reshape $F$ to $\mathbb{R}^{C \times M}$ ($M=H\times W$). Then we perform matrix multiplication between $F$ and the transpose of $F$ and normalize the result to obtain the channel semantic relation map $C \in \mathbb{R}^{C \times C}$.  Specifically, the semantic similarity between any two channels is calculated as: 
\vspace*{-0.2em}
\begin{equation}
\label{eq7}
C_{mn} = \frac{\exp\left(f_{m}^T f_{n}\right)}{\sum_{l=1}^{C} \exp\left(f_{m}^T f_{l}\right)},
\end{equation}
where $f_{m},f_{n}\in \mathbb{R}^M$ denote the features in the $m^{th}$ and $n^{th}$ channels of $F$ respectively.

The channel features are then aggregated based on the channel relation map in the following \textit{aggregation} operation, where we perform matrix multiplication between $C$ and $F$ to obtain the aggregated feature map $E^C \in \mathbb{R}^{C \times M}$:
\vspace*{-0.2em}
\begin{equation}
\label{eq8}
E^C = C F.
\vspace*{-0.2em}
\end{equation}
$E^C$ is finally reshaped to $\mathbb{R}^{C\times H\times W}$ to maintain the input size.
Note that the resulting feature map aggregates semantically similar features according to input-specific channel relation map $C$. This is complementary to SIA which aggregates features according to spatial relation map.
Similar with SIA, CIA can adaptively adjust the input feature map, helping to boost the feature discriminability.

\begin{figure}[t]
\captionsetup{font={small}}
\centering
   \includegraphics[width=1.0\linewidth]{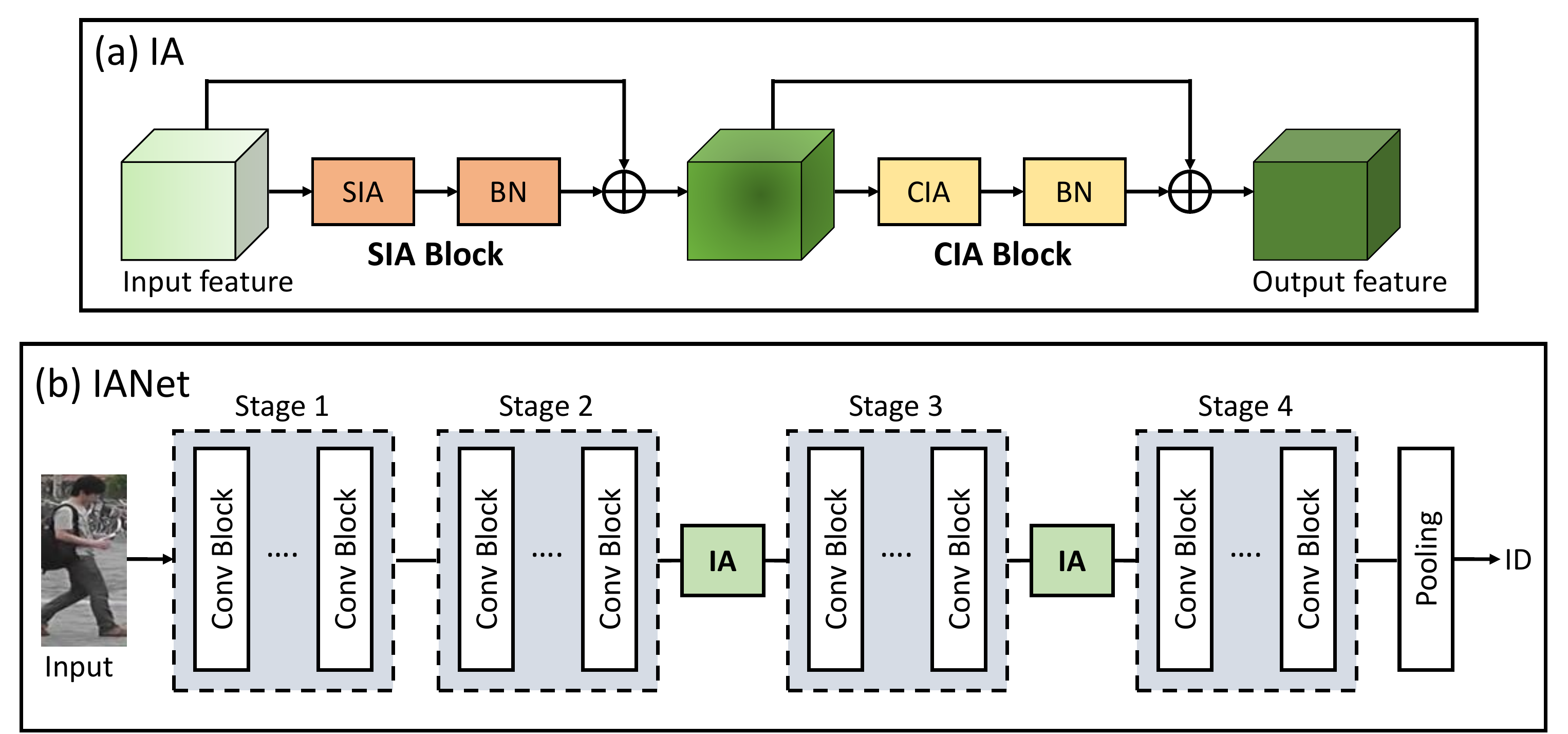}
   \caption{(a) The structure of IA block, which is sequentially consisted of SIA and CIA blocks, (b) The architecture of IANet.}
\label{network}
\vspace*{-0.6em}
\end{figure}

\subsection{IA Block}
We turn the SIA (CIA) module into SIA (CIA) block that can be easily incorporated into existing architectures. As shown in Fig.~\ref{network} (a), SIA (CIA) block is defined as:
\vspace*{-0.2em}
\begin{equation}
\label{eq9}
Y = \text{BN}(E) + F,
\vspace*{-0.2em}
\end{equation}where
$F$ is the input feature map, $E$ is the output of SIA or CIA modules that is given in Eq.~\ref{eq6} or Eq.~\ref{eq8}, and BN is a batch normalization layer~\cite{BN} which adjusts the scale of $E$ with respect to the input. The residual connection ($+F$)~\cite{residual} allows us to insert a new block into any pre-trained model, without breaking its initial performance (\eg the parameters of BN are initialized to zeros).

Given an input feature map, SIA and CIA blocks compute complementary interdependencies. We sequential arrange SIA and CIA blocks to form the IA block (see Fig.~\ref{network} (a)). IA block can be inserted at any depth of a network. Considering the computational complexity, we only place it at the bottlenecks of models where the downsampling of feature maps occurs. Multiple IA blocks located at bottlenecks of different levels can progressively enhance the feature representations with negligible number of parameters.   

\subsection{IANet for Person ReID}

The architecture of IANet is illustrated in Fig.~\ref{network} (b). Here we use ResNet-50~\cite{residual} pre-trained on ImageNet~\cite{imagenet} as the backbone network for person reID. The output dimension of the classification layer is set to the number of training identities. Following~\cite{PCB}, we remove the last spatial down-sampling operation in the backbone network to increase retrieval accuracy with very light computation cost added. IA blocks are then inserted into the backbone network after stage-2 and stage-3 layers. The training procedure of IANet follows the standard identity classification paradigm~\cite{SVdnet,domain-guided,spindle-net}, where the identify of each person is treated as a distinct class. IANet is end-to-end trained with cross-entropy loss. During testing, the features of probe and gallery images are extracted by IANet, and the cosine distance is used for matching.

\subsection{Discussions}
In this subsection, we give a brief discussion on the relations between our proposed IA block and some existing models. 

\textbf{Relations to Non-local} Our IA and Non-local (NL) are both the concrete forms of self-attention. Compared to NL, IA is more suitable to reID because of the following advantages: \textbf{(1)} the proposed CIA is the first attempt to apply self-attention on the channel dimension, which is conductive to highlighting important but small details or body parts. \textbf{(2)} NL can be seen a special case of SIA in the single-context version. Multi-context SIA fuses appearance similarities across multiple patch size, which could localize the body parts more precisely. \textbf{(3)} SIA considers the spatial structure of pedestrians and models the location relations to constrain and complement the appearance similarity.

\textbf{Relations to SCA-CNN and CBAM} SCA-CNN \cite{chen2017sca} and CBAM \cite{woo2018cbam} propose spatial and channel attention to enhance important features and suppress unnecessary ones. However there is no direct guidance for this process, making these methods easily produce unreliable attentions. On the contrary, our IA models generates the attention maps guided by semantic similarity between features, which could adaptively locate body parts and are more reliable.

\textbf{Relations to Squeeze-and-Excitation} CIA has some similarities with Squeeze-and-Excitation network (SE) \cite{SE} because both are designed to model the interdependencies between channels to improve the feature representation power. However, SE computes channel-wise attention that selectively emphasizes informative features, while ignoring the spatial-wise responses due to global spatial pooling. Therefore, the spatial structure information is lost. 

\textbf{Relations to Graph Convolutional Network} SIA and CIA could be treated as the extended Graph Convolutional Network (GCN), where the nodes of graph are defined by the spatial features and channel features, respectively, and the adjacent matrix is the semantic relation map. Compared to conventional GCN where the adjacent matrix is fixed, SIA and CIA change the graph structure adaptively during training, which is more desirable for information propagation between feature nodes.

\section{Experiments}
\label{experiment}
\subsection{Experiment Setup}
\textbf{Datasets and Evaluation Metric.}
We conduct extensive experiments on four person reID benchmarks, CUHK03~\cite{Cuhk}, Market-1501~\cite{Market1501}, DukeMTMC-reID~\cite{Duke} and MSMT17~\cite{msmt17}. For CUHK03, we follow the standard protocol detailed in~\cite{Cuhk} and report the results on manually annotated and DPM-detected images. We adopt mean Average Precision (mAP)~\cite{map} and Cumulative Matching Characteristics (CMC)~\cite{cmc} as evaluation metrics.


\textbf{Implementation details.}
For our implementation, the input images are resized to $256 \times 128$ after random left-right flipping. The initial learning rate is set to $0.0003$. 
Adam~\cite{adam} optimizer is used with a mini-batch size of $32$ for training. The number of context levels ($N$ in Eq.~\ref{eq2}) is set to $3$. 
Because the feature maps at different stages of ResNet have different spatial sizes, we use different standard deviations ($\sigma_{1}$ and $\sigma_2$ in Eq.~\ref{eq3}) for IA blocks at different stages. Specially, $\sigma_{1}$ and $\sigma_2$ are set to $5$ and $10$ when IA blocks are added to stage-3 layers, and $\sigma_{1}$ and $\sigma_2$ are set to $10$ and $20$ when IA blocks are added to stage-2 layers.

\subsection{Comparison with State-of-the-art Approaches}
\begin{table}[t]
\captionsetup{font={small}}
\caption{Comparison with state-of-the-arts on Market-1501 and DukeMTMC. The methods are separated into three groups: global features (\textbf{G}), part features (\textbf{P}) where * denotes those requiring auxiliary part detection, and multi-scale features (\textbf{MS}).}
\small
\centering
\begin{tabular}{c | c |c c |c  c}
\hline
\multicolumn{2}{c|}{\multirow{2}*{Methods}}  & \multicolumn{2}{c|}{Market-1501}  &\multicolumn{2}{c}{DukeMTMC} \\   
\cline{3-6}
\multicolumn{2}{c|}{ }&top-1 &mAP &top-1 &mAP \\
\hline
\multirow{7}*{\textbf{G}} &SVDNet~\cite{SVdnet} &82.3 &62.1 &76.7 &56.8 \\
&MGCAM~\cite{mask-guided} &83.7 &74.3 &-- &-- \\
&BraidNet~\cite{cascaded}  &83.7 &69.5 &76.4 &59.5 \\
&GAN~\cite{Duke}  &84.0 &66.1 &67.7 &47.1 \\
&Adversarial~\cite{adversarially}  &86.4 &70.4 &79.1 &62.1 \\
&Dual~\cite{Interaction-aware}  &91.4 &76.6 &81.8 &64.6 \\
&Mancs~\cite{Mancs}  &93.1 &82.3 &84.9 &71.8 \\
\hline
\multirow{7}*{\textbf{P}} &Spindle*~\cite{spindle-net} &76.9 &-- &-- &-- \\
&PAR~\cite{part-aligned}  &81.0 &63.4 &-- &-- \\
&AACN*~\cite{attention-aware}  & 85.9  &66.9 &76.8 &59.2\\
&PSE*~\cite{pose-sensitive}  &87.7 &69.0 &79.8 &62.0 \\
&HA-CNN~\cite{harmoniou}  &91.2 &75.7 &80.5 &63.8 \\
&SPReID*~\cite{semantic} &92.5 &81.3 &84.4 &70.9 \\
&RPP~\cite{PCB} &93.8 &81.6 & 83.3 &69.2 \\
\hline
\multirow{5}*{\textbf{MS}} &DaRe(R)~\cite{Resource} &86.4 &69.3 & 75.2 &57.4 \\
&DPFL~\cite{multi-scale-representations}  &88.6 &72.6 &79.2 &60.6 \\
&MLFN~\cite{Multi-Level}  &90.0 &74.3 &81.0 &62.8 \\
&KPM~\cite{KPM}  &90.1 &75.3 &80.3 &63.2 \\ 
&Group~\cite{CRF} &93.5 &81.6 &84.9 &69.5 \\
\hline
&IANet &\textbf{94.4} &\textbf{83.1} &\textbf{87.1} &\textbf{73.4} \\
\hline
\end{tabular}
\label{Tab2}
\vspace*{-1em}
\end{table}

\textbf{Market-1501 and DukeMTMC.} 
In Tab.~\ref{Tab2}, we compare IANet with state-of-the-arts on Market-1501 and DukeMTMC. 
IANet achieves the best performance on all evaluation criteria. It is noted that: (1) The gaps between our results and the spatial alignment methods (Spindle~\cite{spindle-net}, AACN~\cite{attention-aware} and PSE~\cite{pose-sensitive}) that incorporate an external part detection sub-network are significant: about $10\%$ improvement on top-1 accuracy and mAP. We argue that these methods are prone to performance degeneration due to inaccurate part detection. On the contrary, our method can adaptively locate the body parts guided by the semantic similarity without external network. (2) IANet outperforms the attention-centric methods (AACN*~\cite{attention-aware}, HA-CNN~\cite{harmoniou} and RPP~\cite{PCB}) that uses spatial attention to learn discriminative parts. We argue that these methods often fail to produce reliable attentions as there is no guidance for this process. On the contrary, our method generates spatial attention maps guided by semantic similarity between spatial features, which are more reliable.  (3) IANet outperforms the multi-scale methods (DaRe~\cite{Resource}, DPFL~\cite{multi-scale-representations}, MLFN~\cite{Multi-Level} and KPM~\cite{KPM}), with an improvement up to $8\%$ on mAP. 
The superiority of IANet over the multi-scale methods indicates that without explicitly fusing features from multiple scales, IANet could also cope with large scale changes.

\begin{table}[t]
   \captionsetup{font={small}}
   \caption{Comparison with state-of-the-art methods on CUHK03.}
   \small
   \centering
   \begin{tabular}{c | c |c c |c  c}
   \hline 
   \multicolumn{2}{c|}{\multirow{2}*{Methods}}  & \multicolumn{2}{c|}{Labeled}  &\multicolumn{2}{c}{Detected} \\   
   \cline{3-6}
   \multicolumn{2}{c|}{ }&top-1 &top-5 &top-1 &top-5 \\ 
   \hline
   \multirow{3}*{\textbf{G}} & SVDNet~\cite{SVdnet} &-- &-- &81.8 &95.2 \\
   &BraidNet~\cite{cascaded}  &88.1 &-- &85.8 &-- \\
   &PN-GAN~\cite{PNGAN} &-- &-- &79.7 &96.2 \\
   \hline
   \multirow{6}*{\textbf{P}} &MSCAN~\cite{context-aware} &74.2 &94.3 &67.9 &91.0 \\
   &PAR~\cite{part-aligned}  &85.4 &97.6 &81.6 &97.3 \\
   &AACN~\cite{attention-aware} &91.3 &98.8 &89.5 &97.6 \\
   &PABR~\cite{part-aligned}  &91.5 &\textbf{99.0} &88.0 &97.6 \\
   &SPReID*~\cite{semantic}  &\textbf{93.8} &98.7 &-- &-- \\
   \hline
   \multirow{5}*{\textbf{MS}}&DPFL~\cite{multi-scale-representations}  &86.7 &-- &82.0 &-- \\
   &CSN~\cite{Multi-Level}  &87.5 &97.8 &86.4 &97.5 \\
   &Group~\cite{CRF} &90.2 &98.5 &88.8 &97.2 \\
   &MLFN~\cite{Multi-Level}  &-- &-- &82.8 &-- \\
   &KPM~\cite{KPM} &91.1 &98.3 &-- &-- \\ 
   \hline
   &IANet &92.4 &\textbf{99.0} &\textbf{90.1} &\textbf{98.2} \\
   \hline
   \end{tabular}
   \label{Tab3}
   \end{table}

\begin{table}[t]
\caption{Comparison with state-of-the-arts on MSMT17.}
\small
\centering
\begin{tabular}{l | c c c c}
\hline
Methods &top-1 &top-5 &top-10 &mAP \\
\hline
GoogleNet~\cite{going} &47.6 &65.0 &71.8 &23.0\\
Pose-driven~\cite{pose-driven} &58.0 &73.6 &79.4 &29.7\\
GLAD~\cite{Glad} &61.4 &76.8 &81.6 &34.0 \\
\hline
IANet &\textbf{75.5} &\textbf{85.5} &\textbf{88.7} &\textbf{46.8} \\
\hline
\end{tabular}
\label{Tab4}
\vspace*{-1.0em}
\end{table}

\textbf{CUHK03.}
In Tab.~\ref{Tab3}, we report the top-1 and top-5 accuracies on CUHK03. 
IAnet outperforms the state-of-the-arts. It is noteworthy that there is a small gap between the labeled evaluation and the detected evaluation of our method, which indicates that our method is robust at the presence of  the imperfect detection. 

\textbf{MSMT17.}
We further evaluate our method on a recent large scale dataset, namely MSMT17. As shown in Tab.~\ref{Tab4}, our method significantly outperform existing works with $14.1\%$ top-1 and $12.8\%$ mAP. Since MSMT17 is the largest dataset with more than $120,000$ images, this result strongly demonstrates the superiority of our proposed method.

\begin{table*}[t]
\caption{Ablations on Market-1501 and DukeMTMC datasets.}
\centering
\small
\subfloat[\textbf{Different Context:} comparison on single-context \textit{interaction} operations of SIA with different patch sizes $K$.]{
\begin{tabular}{l |c c |c  c}
\hline 
\multirow{2}*{$K$}  & \multicolumn{2}{c|}{Market-1501}  &\multicolumn{2}{c}{DukeMTMC} \\   
\cline{2-5}
&top-1 &mAP &top-1 &mAP \\ 
\hline
base. &90.4 &76.2 &82.1 &66.0\\
\hline
$K$=1      &91.8 &\textbf{79.0} &83.6 &\textbf{68.8}\\
$K$=2      &\textbf{93.1} &\textbf{79.0} &\textbf{84.1} &\textbf{68.8}\\
$K$=3      &92.2 &78.9 &83.8 &68.7\\
$K$=5      &91.1 &77.9 &83.5 &67.3\\
\hline
\end{tabular}
\label{taba}
}~~~
\subfloat[\textbf{Different fusion functions $\mathscr{F}$:} comparison on different fusion functions in the multi-context \textit{interaction} operation.]{
\begin{tabular}{l |c c |c  c}
\hline 
\multirow{2}*{$\mathscr{F}$} & \multicolumn{2}{c|}{Market-1501}  &\multicolumn{2}{c}{DukeMTMC} \\   
\cline{2-5}
&top-1 &mAP &top-1 &mAP \\ 
\hline
base. &90.4 &76.2 &82.1 &66.0\\
\hline
 MAX    &93.6  &81.8 &85.2 &71.0 \\
 SUM    &93.5  &81.9 &85.4 &70.9\\
 PROD   &\textbf{93.9}  &\textbf{82.0} &\textbf{85.6} &\textbf{71.5}\\
\hline
\end{tabular}
\label{tabb}
}~~~
\subfloat[\textbf{Location Relation:} comparison on different SIA blocks that respectively use location, appearance and semantic relation maps in the \textit{aggregation} operation.]{
\begin{tabular}{l |c c |c  c}
\hline 
\multirow{2}*{Model} & \multicolumn{2}{c|}{Market-1501}  &\multicolumn{2}{c}{DukeMTMC} \\   
\cline{2-5}
&top-1 &mAP &top-1 &mAP \\ 
\hline
base. &90.4  &76.2 &82.1 &66.0 \\
\hline
Location    &92.1  &79.2 &83.7 &68.5 \\
Appearance  &93.9  &82.0 &85.6 &71.5\\
Semantic &\textbf{94.1}  &\textbf{82.5} &\textbf{85.9} &\textbf{72.2}\\
\hline
\end{tabular}
\label{tabc}
}

\subfloat[\textbf{Combining methods of CIA and SIA blocks.}]{
\begin{tabular}{L{1.5cm} |C{0.7cm} C{0.5cm} |C{0.7cm}  C{0.4cm}}
\hline 
\multirow{2}*{Combine} & \multicolumn{2}{c|}{Market-1501}  &\multicolumn{2}{c}{DukeMTMC} \\  
\cline{2-5}
&top-1 &mAP &top-1 &mAP \\ 
\hline
base.       &90.4  &76.2 &82.1 &66.0\\
CIA            &91.9 &79.3 &84.3 &68.7 \\
\hline
SIA \& CIA    &94.0  &82.5 &85.7 &72.2\\
CIA + SIA      &94.1  &82.5 &85.7 &\textbf{72.3}\\
SIA + CIA      &\textbf{94.3}  &\textbf{82.8} &\textbf{85.9} &\textbf{72.3}\\
\hline
\end{tabular}
\label{tabd}
}~~~
\subfloat[\textbf{Positions to place IA blocks:} an IA block is added into the bottlenecks of different stages.]{
\begin{tabular}{L{0.7cm} |C{0.7cm} C{0.5cm} |C{0.7cm}  C{0.4cm}}
\hline 
\multirow{2}*{Model} & \multicolumn{2}{c|}{Market-1501}  &\multicolumn{2}{c}{DukeMTMC} \\   
\cline{2-5}
&top-1 &mAP &top-1 &mAP \\ 
\hline
base.       &90.4  &76.2 &82.1 &66.0\\
\hline
stage$_{1}$ &93.7  &81.8 &85.3 &71.5\\
stage$_{2}$ &\textbf{94.4}  &\textbf{82.8} &\textbf{86.5} &71.8\\
stage$_3$ &94.3  &82.8 &85.9 &\textbf{72.3}\\
stage$_4$ &92.3  &79.6 &84.8 &69.3\\
\hline
\end{tabular}
\label{tabe}
}~~~
\subfloat[\textbf{Bottleneck \textit{vs.} Inside each Convolution Block:} stage$_{23}$-c denotes IA blocks are inserted to each convolution block in stage$_2$ and stage$_3$ layers.]{
\begin{tabular}{L{1.2cm} |C{0.7cm} C{0.5cm} |C{0.7cm}  C{0.4cm}}
\hline 
\multirow{2}*{Model} & \multicolumn{2}{c|}{Market-1501}  &\multicolumn{2}{c}{DukeMTMC} \\   
\cline{2-5}
&top-1 &mAP &top-1 &mAP \\ 
\hline
base.       &90.4  &76.2 &82.1 &66.0\\
stage$_{2}$ &94.4  &82.8 &86.5 &71.8\\
stage$_3$ &94.3  &82.8 &85.9 &72.3\\
\hline
stage$_{23}$-c &93.6 &81.9 &86.1 &72.2\\
stage$_{23}$ &\textbf{94.4} &\textbf{83.1} &\textbf{87.1} &\textbf{73.4} \\
\hline
\end{tabular}
\label{tabf}
}
\vspace*{-1em}
\end{table*}

\subsection{Ablation Study}
In this section, we investigate the effectiveness of each component in IA block by conducting a series of ablation studies on Market-1501 and DukeMTMC datasets. We adopt ResNet-50~\cite{residual} trained with cross-entropy loss as the baseline (denoted as base.). If there is no special explanation, we add the proposed blocks to the last residual block (bottleneck) of $\text{stage}_{3}$ layer of ResNet-50. 

\textbf{Multi-context combination.}
We first compare the single-context \textit{interaction} operations ($S^A_K$ in Eq.~\ref{eq1}) with different patch sizes $K$. In this part, SIA uses single-context appearance relation map in the \textit{aggregation} operation. As shown in Tab.~\ref{taba}, there is an improvement in performance when $K$ is increased, showing the effectiveness of incorporating contextual information. However, as $K$ is further increased, the accuracy drops gradually. 
So we only use context patches with $K$ of $1$, $2$ and $3$ in the multi-context \textit{interaction} operation. We then explore three different fusion functions $\mathscr{F}$ in the multi-context \textit{interaction} operation ($S^A$ in Eq.~\ref{eq2}): element-wise maximum, summation, and product. In this part, SIA uses multi-context appearance relation map in the \textit{aggregation} operation. Tab.~\ref{tabb} lists the comparison results of different fusion strategies. Element-wise product performs better than other functions and is therefore selected as the default fusion function. 

We finally report the computation cost of SIA. For single-context SIA ($K$=1), the relation map $S^{A}_{1}$ could be worked out by one matrix multiplication. The multi-context SIA ($K$=1,2,3) does not incur extra multiplying operation compared to single-context, as $S^{A}_{1}$ has got the dot-product between every two spatial position. Specially, baseline requires $\sim$4.06 Multiply GFLOPs in a single forward pass for a 256$\times$128 pixel input image. Baseline added SIA requires $\sim$4.09 Multiply GFLOPS, corresponding only $\sim0.73\%$ relative increase over ResNet50.

\begin{figure}[t]
\captionsetup{font={small}}
\small
\centering
   \includegraphics[width=1.0\linewidth]{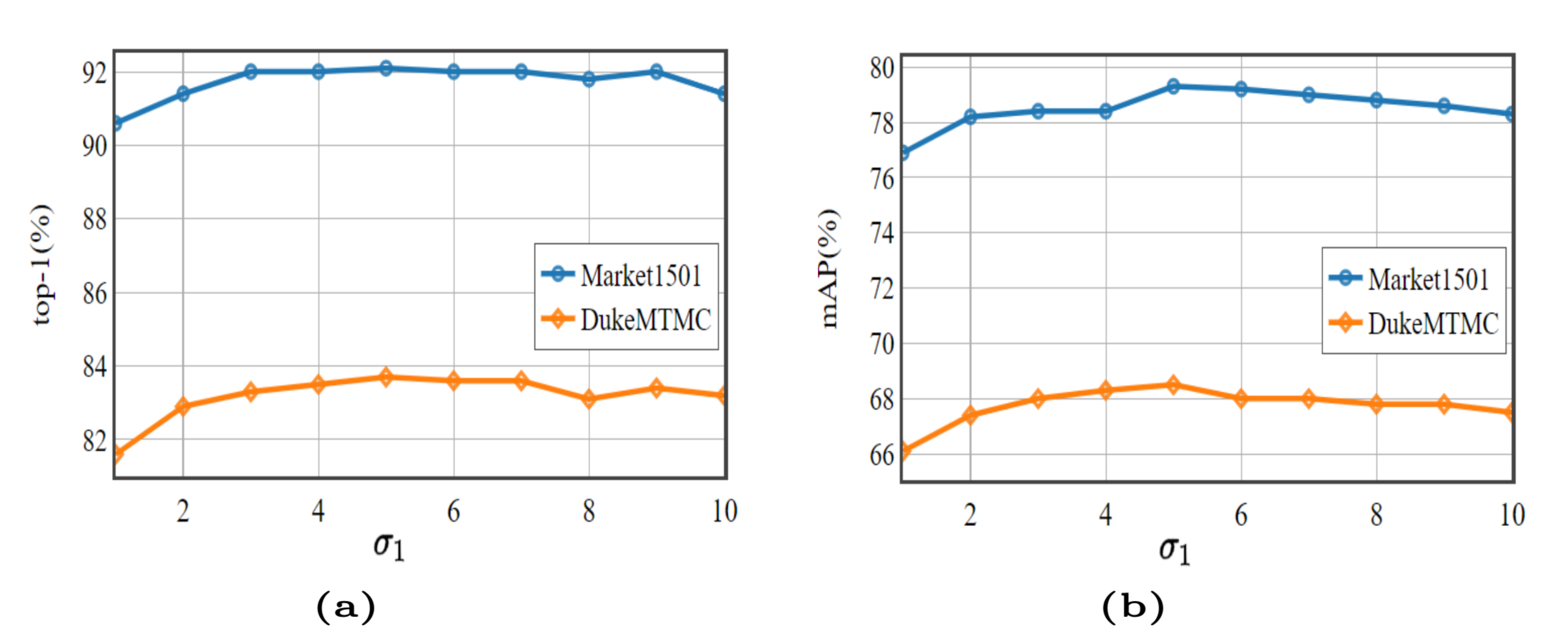}
   \vspace*{-2em}
   \caption{Parameter analysis for location relation map. (a) top-1 accuracy changes with the standard deviation $\sigma_1$. (b) mAP changes with the standard deviation $\sigma_1$.}
\vspace*{-2em}
\label{location}
\end{figure}

\textbf{Effectiveness of introducing location prior.}
In the location relation map, the standard deviations $\sigma_1$ and $\sigma_2$ determine the shape of Gaussian function. In order to reduce the search space of hyper-parameters, we set $\sigma_2$ to $2\sigma_1$ according to the input aspect ratio $2$:$1$. Fig.~\ref{location} shows the top-1 accuracy and mAP changes with $\sigma_1$, where SIA uses spatial location relation map in the \textit{aggregation} operation. We can see that the performance of SIA is not sensitive to $\sigma_1$ with a certain range of values. 

We then investigate whether location relations can improve the performance of SIA. Tab.~\ref{tabc} compares different SIA blocks: Location, Appearance and Semantic that respectively use location ($S^L$), appearance ($S^A$) and semantic ($S$) relation maps in \textit{aggregation} operation. As seen, Location achieves better results than baseline, which demonstrates the effectiveness of introducing location relations. Besides, Semantic consistently outperforms Appearance, showing that location relations complement appearance relations, which can achieve more precise semantic relations.  

\textbf{Arrangement of CIA and SIA blocks.} 
We first verify the effectiveness of CIA block in Tab.~\ref{tabd} by adding it to baseline. 
CIA outperforms baseline by $\sim 3\%$ in mAP, which implies that it is effective to enhance feature representation power by aggregating similar channel features. 
We then compare three different ways of arranging CIA and SIA blocks: parallel, sequential channel-spatial, and sequential spatial-channel. As shown in Tab.~\ref{tabd}, the sequential spatial-channel produces the best performance. Note that the results outperform adding CIA or SIA blocks independently, showing that utilizing both blocks is crucial and the best-arranging strategy further pushes performance. 

\textbf{Efficient positions to place IA blocks.}
Table \ref{tabe} compares an IA block added to the bottlenecks of different stages of ResNet. The improvements of an IA block in stage$_2$ and stage$_3$ are similar, but smaller in stage$_1$ and stage$_4$. 
Therefore, we only insert IA blocks into stage$_2$ and stage$_3$ layers. 
Finally, we empirically verify that the bottlenecks are the effective positions to place IA blocks. Recent studies~\cite{non-local,SE,residual-attention} mainly focus on modifications within the '\textit{convolution blocks}' rather than the '\textit{bottlenecks}'. Tab.~\ref{tabf} compares two different locations, where stage$_{23}$ adds $2$ IA blocks to the bottlenecks, while stage$_{23}$-c adds $10$ blocks to every residual block of stage$_2$ and stage$_3$ layers. We can clearly observe that placing the blocks at the bottlenecks is more effective. We argue that too many IA blocks are difficult to optimize on small reID datasets.

\begin{table}[t]
\begin{small}
\caption{Performance \wrt different backbones on Market-1501 dataset.}
\vspace*{-1.8em}
\label{tab2}
\begin{center}
\begin{tabular}{c | c c |c c | c c}
\hline
model & R32 & R32-IA & R101 & R101-IA & GN &GN-IA\\
\hline
top-1 & 88.7& 92.6& 91.1 &94.0 &83.4 &87.1  \\
mAP &72.5 &79.2 &76.5 &83.0 &64.0 &69.7\\
\hline
\end{tabular}
\end{center}
\end{small}
\vspace*{-1.7em}
\end{table}

\textbf{Effectiveness of IA block across different backbones.} In order to further verify the validity of IA block, we try another three backbones besides the ResNet-50, \ie, ResNet32 (R32), ResNet101 (R101), and GoogleNet (GN). As shown in Tab.~\ref{tab2}, our method (-IA) improves the performance \wrt different backbones consistently.

\begin{figure}[t]
\captionsetup{font={small}}
\centering
   \includegraphics[width=0.8\linewidth]{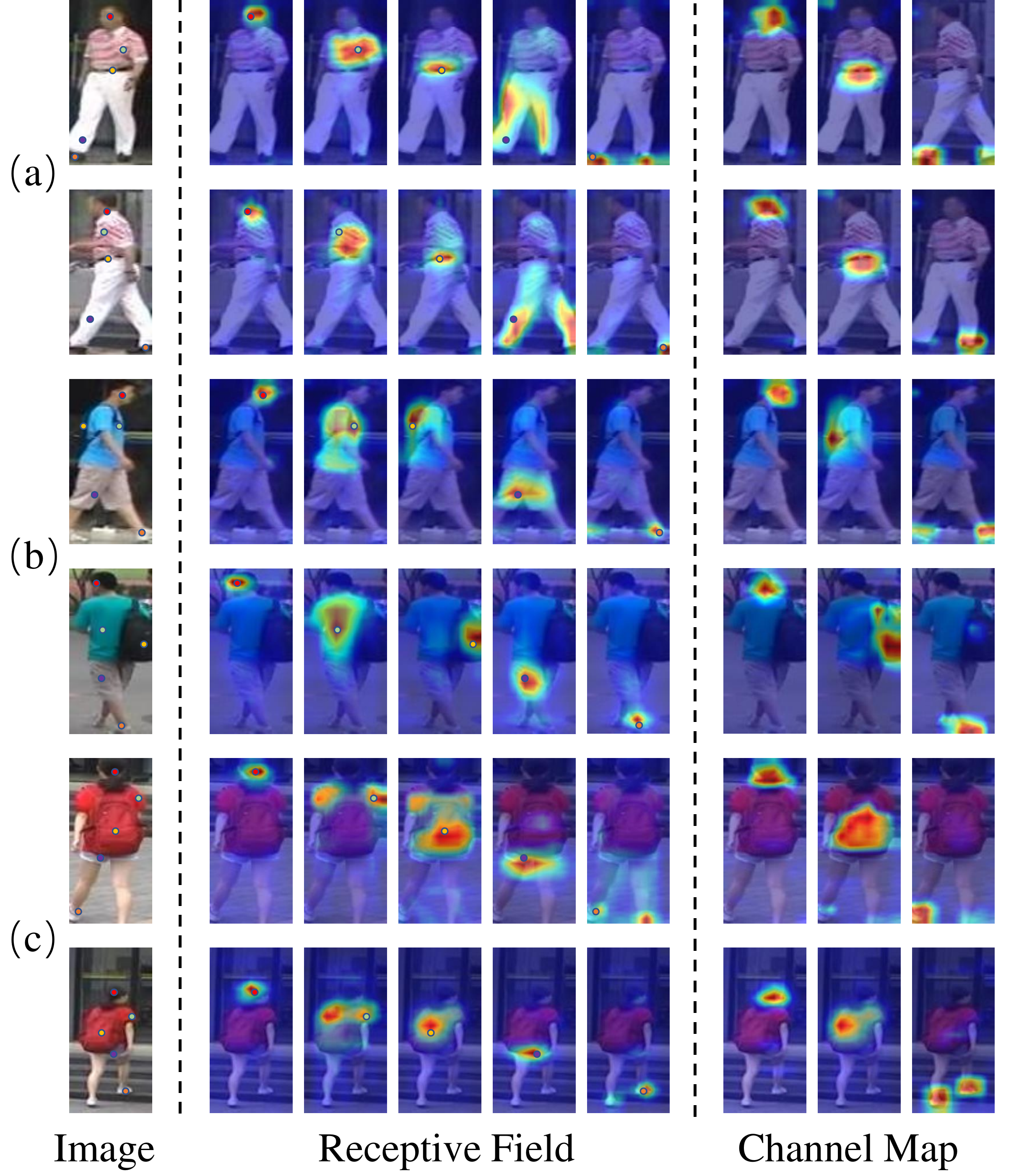}
   \caption{Visualization results of SIA and CIA on Market-1501. For each row, we show an input image, five receptive fields corresponding to the points masked in the input image, and three channel maps from the output of CIA. The channel maps are from $5^{th}$, $11^{th}$ (a) / $25^{th}$ (b, c) and $121^{th}$ channels, respectively.}
\label{vis}
\vspace*{-1.3em}
\end{figure}

\subsection{Visualization for Pose and Scale Robustness}
To verify whether SIA can adaptively localize body parts under various poses and scales, we visualize the pixel-wise receptive fields learned with SIA. Specifically, 
each specific position has a corresponding sub-relative map. 
We define the sub-relation maps with high relation values as valid receptive fields and highlight them. As shown in Fig.~\ref{vis}, for each input image, we select five positions from head, torso, belt/bag, legs and shoe and show their corresponding valid receptive fields. We can clearly observe that SIA can adaptively localize the body parts and visual attributes under various poses and scales. For example, in Fig.~\ref{vis} (a) and Fig.~\ref{vis}  (b), the person take on different poses. SIA could aggregate the features of regions corresponding to the body parts independently of the pose. In Fig. \ref{vis} (c), the person is in different scales due to the detection errors. SIA could adaptively adjust the scales of receptive fields based on the scales of body parts. In addition, the receptive fields of different body parts in SIA have different shapes and scales, which is superior to the fixed geometric receptive field in CNNs.

For CIA, it is hard to give comprehensive visualization about the relation maps directly. Instead, we show some aggregated channels to see whether they highlight small body parts or attribute areas. In Fig. \ref{vis}, we display $5^{th}$, $11^{th}$ (a) / $25^{th}$ (b, c) and $121^{th}$ channels. We find that the responses of specific body parts and attributes are noticeable after CIA enhances. For example, $11^{th}$, $25^{th}$ and $121^{th}$ channel maps respond to attribute belt, bag and shoes. In short, the visualizations further demonstrate the necessity of modeling the interdependencies between channels for improving feature representation, especially for fine-grained attributes. 

\section{Conclusion}
In this paper, we propose SIA and CIA blocks to improve the representational capacity of deep convolutional networks. SIA models the interdependencies between spatial features of convolutional feature maps. It can adaptively localize the body parts under various poses and scales. CIA models the interdependencies between channel features. It can further enhance the feature representations especially for small visual cues. Extensive experiments show that IANet outperforms state-of-the-arts on three public person reID datasets.   

\noindent\textbf{Acknowledgement} 
This work is partially supported by National Key R\&D Program of China (No.2017YFA0700800), Natural Science Foundation of
China (NSFC): 61876171 and 61572465. 

{\small
\bibliographystyle{ieee_fullname}
\bibliography{egbib}
}

\end{document}